\documentclass[10pt,twocolumn,letterpaper]{article}

\usepackage[pagenumbers]{cvpr} %

\usepackage[T1]{fontenc}

\usepackage{xcolor}
\definecolor{mediumgray}{rgb}{0.5,0.5,0.5}
\newcommand{\pmsd}[1]{{\color{mediumgray}{\tiny{ $\pm$ #1}}}}

\usepackage{graphicx}
\usepackage{booktabs}
\usepackage{multirow}
\usepackage{url}

\usepackage{array}
\usepackage{mdframed}

\newcommand{\boldparagraph}[1]{\vspace{0.2cm}\noindent{\textbf{#1:}} } %

\newcommand{\insight}[1]{
\begin{mdframed}[leftmargin=.5\parindent, rightmargin=.5\parindent] \centering \it
\textit{#1}
\end{mdframed}
}%

\definecolor{darkgreen}{rgb}{0,0.7,0}
\definecolor{darkblue}{RGB}{31,119,180}
\definecolor{darkred}{RGB}{214,39,40}

\newcommand{\legend}[6]{
    \centerline{
    \ifbool{#1}{\raisebox{-1px}{\includegraphics[height=8px]{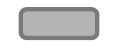}}\footnotesize{Ego}\hspace{2px}}{}
    \ifbool{#2}{\raisebox{-1px}{\includegraphics[height=8px]{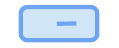}}\footnotesize{Vehicle}\hspace{2px}}{}
    \ifbool{#3}{\raisebox{-1px}{\includegraphics[trim={5mm 0mm 5mm 0mm}, clip, height=8px]{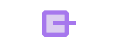}}\footnotesize{Pedestrian}\hspace{2px}}{}
    \ifbool{#4}{\raisebox{-1px}{\includegraphics[trim={3mm 0 3mm 0}, clip, height=8px]{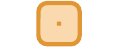}}\footnotesize{Static}\hspace{2px}}{}
    \ifbool{#5}{\raisebox{-1px}{\includegraphics[trim={6mm 1mm 6mm 1mm}, clip, height=8px]{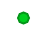}}\footnotesize{Waypoint}\hspace{2px}}{}
    \ifbool{#6}{\raisebox{-1px}{\includegraphics[trim={6mm 1mm 6mm 1mm}, clip, height=8px]{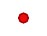}}\footnotesize{Path}\hspace{2px}}{}
    }
}

\definecolor{cvprblue}{rgb}{0.21,0.49,0.74}
\usepackage[pagebackref,breaklinks,colorlinks,allcolors=cvprblue]{hyperref}

\def\paperID{*****} %
\def\confName{CVPR}
\def\confYear{2026}

\title{PlanT 2.0: Exposing Biases and Structural Flaws in Closed-Loop Driving}

\author{
Simon Gerstenecker \qquad
Andreas Geiger \qquad
Katrin Renz \\
[4mm]
University of Tübingen \quad
Tübingen AI Center \quad \\
}

\begin{document}
\maketitle
\setlength\parindent{0pt}
\begin{abstract} Most recent work in autonomous driving has prioritized benchmark performance and methodological innovation over in-depth analysis of model failures, biases, and shortcut learning. This has led to incremental improvements without a deep understanding of the current failures. While it is straightforward to look at situations where the model fails, it is hard to understand the underlying reason. This motivates us to conduct a systematic study, where inputs to the model are perturbed and the predictions observed.
We introduce PlanT 2.0, a lightweight, object-centric planning transformer designed for autonomous driving research in CARLA. The object-level representation enables controlled analysis, as the input can be easily perturbed (e.g., by changing the location or adding or removing certain objects), in contrast to sensor-based models. To tackle the scenarios newly introduced by the challenging CARLA Leaderboard 2.0, we introduce multiple upgrades to PlanT, achieving state-of-the-art performance on Longest6 v2, Bench2Drive, and the CARLA validation routes.
Our analysis exposes insightful failures, such as a lack of scene understanding caused by low obstacle diversity, rigid expert behaviors leading to exploitable shortcuts, and overfitting to a fixed set of expert trajectories. Based on these findings, we argue for a shift toward data-centric development, with a focus on richer, more robust, and less biased datasets.

We open-source our code and model at \url{https://github.com/autonomousvision/plant2}.

\end{abstract}    
\section{Introduction}
\label{sec:intro}
Autonomous driving (AD) remains an unsolved challenge despite decades of research~\cite{Pomerleau1988NIPS, LeCun2004DAVE, Codevilla2018ECCV, Chitta2023PAMI}. Two of the key missing abilities are robustness and generalization to unseen situations. 
In academic research, the CARLA simulator~\cite{CarlaDosovitsky} has become the de facto standard for developing and evaluating driving models in a closed-loop manner~\cite{Wu2022NeurIPS,Chitta2023PAMI, Shao2022CORL}. 
While other benchmarks, such as nuPlan~\cite{Caesar2021CVPRW} or NAVSIM~\cite{Dauner2024NEURIPS}, offer highly realistic sensor inputs, the CARLA simulator is particularly valued for its high degree of customizability, enabling controlled experiments with scenarios that would be dangerous or extremely rare in the real world.

The first CARLA Leaderboard (v1.0) helped to unify evaluation but had many shortcomings: scenarios were overly simple and not diverse, slow driving speeds were not penalized, pedestrians became irrelevant due to the slow driving speed, and stop signs could be fully ignored~\cite{Chitta2023PAMI, Renz2022CORL}. The newer Leaderboard 2.0 aimed to overcome these issues by enforcing higher driving speeds and incorporating a broader set of complex scenarios~\cite{Leaderboard2024}. However, progress on this leaderboard has been slow due to the newly introduced complexity~\cite{zimmerlin2024tfpp,Renz2025cvpr}. We argue that a systematic investigation of failures, biases, and shortcuts can help to make non-trivial advancements in the field. 
However, for end-to-end models, it is time-consuming and non-trivial to perturb sensor inputs such as RGB images or LiDAR point clouds in a controlled fashion, making it difficult to isolate and study specific behaviors of end-to-end sensor-based models. 
As a result, systematic investigations of bias and structural flaws remain rare.

To address these gaps, we revisit PlanT~\cite{Renz2022CORL}, a lightweight privileged planner with an object-centric input representation. We present \textbf{PlanT 2.0}, an extension of the original model with richer input modalities and improved output representation, achieving state-of-the-art performance across multiple CARLA 2.0 benchmarks.
PlanT is particularly well-suited for in-depth analysis for two reasons:
(1) its object-level inputs make scene perturbations straightforward, enabling controlled studies such as shifting vehicles, adding pedestrians, or modifying traffic signals, and
(2) its computational efficiency allows for fast training and evaluation.

In this work, we go beyond performance metrics: Through extensive controlled experiments, we uncover biases, structural flaws, and shortcut learning. Specifically, we find: (1) limited obstacle diversity, as present in the default CARLA Leaderboard 2.0 scenarios,  prevents true scene understanding, (2) fixed expert trajectories cause overfitting and poor generalization, and (3) shortcut learning emerges from overly deterministic expert demonstrations provided by the privileged expert used to collect training data for the current state-of-the-art models. 
Our analysis demonstrates potential for major improvements by focusing on richer, more robust, and less biased datasets for training, suggesting a necessary shift in perspective.

\section{Related Work}

\boldparagraph{Closed-Loop Driving in CARLA}
The CARLA simulator~\cite{CarlaDosovitsky} has gained significant popularity and has become the most established self-driving simulator. It enables closed-loop evaluation of driving policies, providing pseudo-realistic sensor outputs for camera and lidar modalities. \\
 Most works utilizing CARLA focus on sensor-based driving. A wide range of approaches have been developed for this task, including vision-only approaches~\cite{Codevilla2018ICRA,Hu2022NeurIPS,Wu2022NeurIPS,Chen2019CORL,Chen2021ICCVa,hu2023_uniad,jia2025drivetransformer,tang2025hipadhierarchicalmultigranularityplanning}, sensor fusion models~\cite{Jia2023CVPR,Chitta2023PAMI,zimmerlin2024tfpp,Shao2022CORL}, and language models~\cite{sima2023drivelm,wang2023drivemlm,shao2023lmdrive,Renz2025cvpr,Fu2025Orion}. \\
The task of privileged planning in CARLA has also attracted considerable attention over the years~\cite{Chen2019CORL,Renz2022CORL,Zhang2021ICCV,li2024think2drive,Beißwenger2024PdmLite,JaegerCarl2025}, particularly following the release of the CARLA leaderboard 2.0~\cite{Leaderboard2024}, which did not include a public expert policy. The rule-based expert PDM-Lite~\cite{Beißwenger2024PdmLite} is a popular choice for imitation learning due to its strong performance and computational efficiency. Another notable method, Think2Drive~\cite{li2024think2drive}, employs reinforcement learning but is limited to the Bench2Drive dataset, since the model was not released publicly. 
Our model, PlanT~2.0, uses PDM-Lite as an expert policy during training, but does not rely on handcrafted, scenario-specific rules or heuristics. In contrast to Think2Drive, PlanT~2.0 is based on imitation learning and uses a sparse representation for other traffic participants, allowing individual treatment of objects independent of distance. Additionally, it requires far less training resources and is fully open-source. 

\boldparagraph{In-Depth Failure Analysis}
Most of the mentioned works improve upon some benchmark and offer novel approaches, but do not clearly investigate the capabilities and remaining failures of the proposed approaches. 
Some works~\cite{Shao2022CORL,Jia2023CVPR,Jia2023ICCV,Wu2022NeurIPS,Renz2025cvpr} note some of the failures observed during evaluation. Based on a qualitative scenario-based analysis of failure modes, Zimmerlin et al.~\cite{zimmerlin2024tfpp} find some issues with the dataset and expert design. However, in-depth analysis of failure causes and model biases is difficult to achieve for sensor-based methods, as the perception module introduces an additional layer of uncertainty about the root cause of the failure in addition to high computational requirements. 

Many works focus on failures of the perception module in isolation by injecting noise or performing adversarial attacks~\cite{wu2020physicaladversarialattackvehicle,nesti2021evaluatingrobustnesssemanticsegmentation,Suryanto2022cvprDTA,wang2021CVPRattentionsuppressionattack}. However, making observations about the planning behavior of a full autonomous driving system is not trivial. Some works adapt the same approaches of injecting faults or adversarial attacks for testing driving models~\cite{zhang2022adversarialrobustnesstrajectoryprediction,maleki2023carfase, piazzesi2022attacksfaultsinjectionselfdriving, jia2025drivetransformer}, while another range of approaches implement automatic or even adversarial scenario generation methods to find novel failure cases~\cite{ramakrishna2022anticarlaadversarialtestingframework,lu2024realisticcornercasegeneration,ramakrishna2022riskawarescenesamplingdynamic,osiński2021carlarealtrafficscenarios}. 
Most of these works only provide methods to induce failures in driving models and do not investigate how these failures came to be. Instead, our work focuses on analyzing the weaknesses of our model, PlanT~2.0, and generating actionable insights for improvement.

\section{Method} %

In order to perform a detailed analysis of model behaviors on the CARLA Leaderboard 2.0~\cite{Leaderboard2024},
we upgrade PlanT~\cite{Renz2022CORL}, as it enables easy manipulation of the object-based inputs for analysis and requires significantly less computational resources compared to end-to-end models.

PlanT~\cite{Renz2022CORL} is a planning transformer designed for the task of vehicle motion planning in the CARLA simulator~\cite{CarlaDosovitsky}. It relies on ground-truth perception as input. While PlanT achieved expert-level performance on the CARLA Leaderboard 1.0, its general capabilities are limited and tightly coupled to the benchmark. For instance, it ignores pedestrians, fails to respect stop signs, and drives at unnaturally low speeds. Moreover, Leaderboard 1.0 itself only covers a restricted set of simple scenarios. In contrast, the newer Leaderboard 2.0 enforces minimum driving speeds and includes a much broader range of complex situations. \\
To cover these new challenges and move closer towards real-world driving, we introduce multiple improvements to PlanT and introduce \textbf{PlanT 2.0}.

\boldparagraph{Input representation}
The object representation is extended to include five new object classes: pedestrians, static objects, emergency vehicles, stop signs, and traffic lights.
Pedestrians and static objects were irrelevant for driving in CARLA leaderboard 1.0 and not part of the input. %
To allow the model to react differently to emergency vehicles, we add a separate object class for emergency vehicles. Lastly, the previous version of PlanT only used a binary flag in the waypoint GRU to represent a red traffic light. We introduce traffic lights and stop signs as bounding boxes representing the stopping line, allowing the model to come to a stop in the correct position. 
Given the increased number of object classes, we use a separate linear projection for each class.\\
Additionally, we add information about the road layout by inputting a simple BEV rendering of the surrounding 64m of road network at a resolution of 0.5m, encoding it to a single token using a ResNet-18. 

\boldparagraph{Input range}
To account for faster driving speeds and higher scenario difficulty, we increase the object detection range from the previous radius of 30m. We use a circular radius of 50m for objects behind the ego vehicle, and an ellipse with an extent of 100m in the x-axis and 50m in the y-axis for objects in front of the vehicle.

\boldparagraph{Output representation}
Finally, we upgrade the planning head of PlanT. Instead of using waypoints for both lateral and longitudinal planning, we use spatially equidistant path points for lateral planning, as well as traditional temporal waypoints for longitudinal planning. This approach was shown to reduce collisions by improving lateral control \cite{Jaeger2023ICCV,Renz2025cvpr}. %
Instead of predicting these points using a GRU on a single token encoded by the transformer backbone, we implement the approach used by \cite{Renz2025cvpr}, which encodes a learned token per point, projects them to two dimensions using two separate linear layers for waypoints and path-points, and cumulatively sums the points for each mode. To account for the changed driving dynamics due to higher speeds, we replace the lateral and longitudinal controllers with those used by the expert PDM-Lite\cite{Beißwenger2024PdmLite}. We compare different output representations and generation methods that are commonly used in the literature, but have never been compared in a unified setting. The details about the settings we compare can be found in Appendix~\ref{supp:model}.

\section{Experiments}

\subsection{Implementation Details}
The architecture of PlanT 2.0 (e.g., number of layers, feature dimensions) is taken from the original PlanT work.
We train our model for 30 epochs using a batch size of 128 with a learning rate of 1e-4. For the final epoch, we reduce the learning rate by a factor of 10. Training on two GeForce RTX 2080 Ti GPUs takes around six hours. We train our model and all ablations using three seeds and evaluate it using three evaluation seeds. Standard deviations are reported across training seeds. 
We base our dataset on the one used by \cite{zimmerlin2024tfpp}, which contains around 500k samples, roughly equivalent to the "2x" dataset used by PlanT\cite{Renz2022CORL}. The dataset is generated using the PDM-Lite expert\cite{Beißwenger2024PdmLite}, a rule-based system that is able to solve all scenarios of the CARLA Leaderboard 2.0. 
We include Town13 in this dataset and perform an ablation without Town13 in Table~\ref{tab:results_validation}. In addition, we introduce small modification to the expert and dataset, which can be found in Appendix~\ref{supp:implementation}.

\subsection{Benchmarks}
We compare the performance of our model on three different commonly used benchmarks.

First, the \textit{official CARLA 2.0 validation routes}, which consist of 20 long routes provided by the CARLA team and include 36 different scenarios. The average length of these routes is 12.4 kilometers. 

\textit{Longest6 v2} consists of 36 medium-length routes, around 1.5km on average, which were originally designed for the CARLA leaderboard 1.0, meaning they do not include the challenging new scenarios and test simpler driving behaviors. The routes were converted to the current version by \cite{zimmerlin2024tfpp}. 

\textit{Bench2Drive}~\cite{jia2024bench} is a widely used benchmark that aims to reduce the computational burden of evaluations by providing five very short (150m) routes for 44 scenarios, totaling 220. These short routes are designed for rapid evaluation and investigation of model performance per scenario, utilizing a balanced distribution of scenarios, as opposed to the validation routes. 
Since PlanT does not use perception inputs, the computational burden of evaluations is significantly reduced, allowing us to perform the main ablations on the official validation routes.
Details about the metrics for each benchmark are described in Appendix~\ref{supp:eval_metrics}.\

\subsection{Evaluation Metrics}
\label{supp:eval_metrics}
The main metric in CARLA is the Driving Score (DS), which represents the multiplication of the Route Completion (RC) and Infraction Score (IS). \cite{zimmerlin2024tfpp} note that since the infraction penalties are multiplied to obtain the IS, the resulting DS does not behave linearly. With increasing Route Completion and a constant infraction rate, the resulting DS may even decrease, an effect which is especially apparent on long routes. They propose the Normalized Driving Score (NDS), which does not suffer from this effect. We report the NDS for experiments on the CARLA validation routes, as they are very long and usually incur multiple infractions, leading to inconsistent Driving Scores. For Longest6 v2 and Bench2Drive, we report the official Driving Score. 

Another interesting metric that can be obtained from Bench2Drive is the Success Rate (SR) across scenarios, which measures the ratio of correctly solved scenarios on Bench2Drive and is not influenced by the severity of failures.

\subsection{State-of-the-Art Performance}

Before conducting any in-depth analysis or ablations, we first aim to establish a state-of-the-art baseline. Our goal is to verify that our model operates within a strong performance regime where failure cases are not dominated by trivial issues. To ensure that subsequent analyses are meaningful, we therefore benchmark PlanT 2.0 against state-of-the-art methods while maintaining the ability to manipulate inputs through an object-centric representation. This allows us to both evaluate competitiveness and perform controlled interventions later on.

As shown in Table~\ref{tab:results_validation}, our model achieves a normalized driving score of 28.6 on the official validation routes. While we fall short of the expert's performance, we reach similar route completion values. Since our training dataset includes samples of the validation town, we train an additional model, excluding samples from Town13 (same settings as~\cite{zimmerlin2024tfpp}). This model exhibits a significant drop in performance of more than 50\%, obtaining a normalized driving score of 13.4. This performance gap likely results from the model greatly benefiting from training directly on Town13, reducing the gap between training and evaluation. Additionally, the drop in performance could stem from the significantly reduced diversity in the dataset. Almost 40\% of our original dataset comprises Town13 routes. We upsample the remaining part of the dataset to match the original dataset scale, which leads to reduced diversity. %

\begin{table}[t]
    \centering
    \setlength{\tabcolsep}{0.015\textwidth}
        \begin{tabular}{c|l|ccc}
            \toprule
            & \textbf{Method} &
            \multicolumn{1}{c}{\textbf{NDS} $\uparrow$} &
            \multicolumn{1}{c}{\textbf{RC} $\uparrow$} \\
            \midrule
            \multirow{3}{*}{\textbf{S}}
            & UniAD~\cite{hu2023_uniad} & 0.00 & 1.42 \\
            & TF++~\cite{zimmerlin2024tfpp}& 4.94 & 68.53 \\
            & TF++ w/o Town13 & 2.12 & 50.20 \\
            \midrule
            \multirow{2}{*}{\textbf{P}} & PlanT 2.0 & 
            28.6\pmsd{2.9} & 
            90.6\pmsd{3.9}
            \\
            & PlanT 2.0 w/o Town13& 
            13.4 \pmsd{2.9} & 83.4 \pmsd{2.7}
            \\
            \midrule
            \textit{P} & \textit{Expert~\cite{sima2023drivelm}}&
            \textit{61.55}&
            \textit{92.35}
            \\
            \bottomrule
        \end{tabular}
    
    \caption{\textbf{CARLA validation routes results.} \textbf{S} denotes sensor based models, \textbf{P} denotes privileged models. The scores of the expert are taken from \cite{zimmerlin2024tfpp}. We report only models where NDS on the CARLA validation routes were publicly available.} 
    \label{tab:results_validation}
\end{table}

We additionally evaluate the model on Bench2Drive, a popular benchmark that includes small routes containing the new CARLA scenarios. As shown in Table~\ref{tab:results_bench2drive}, our model outperforms the driving score reported for the privileged, learning-based Think2Drive model, setting a new state of the art for learning based planners. On Bench2Drive, the difference in performance between our models and the expert is not as large as on the validation routes, but a performance gap is still present. A large number of failures result from obstacle avoidance scenarios, such as ConstructionObstacleTwoWays, Accident, and ParkedObstacle. We will investigate such failures in further detail in the following chapter.

\begin{table}[t]
    \centering
    \setlength{\tabcolsep}{0.015\textwidth}
        \begin{tabular}{c|l|>{\centering\arraybackslash}p{1.5cm}>{\centering\arraybackslash}p{1.5cm}}
            \toprule
            & \textbf{Method} &
            \multicolumn{1}{c}{\textbf{DS} $\uparrow$} &
            \multicolumn{1}{c}{ \textbf{SR} $\uparrow$}\\
            \midrule
            \multirow{4}{*}{\textbf{S}}
            & Raw2Drive~\cite{yang2025raw2drivereinforcementlearningaligned} & 71.36 & 50.24 \\
            & TF++~\cite{zimmerlin2024tfpp}& 84.21 & 67.27 \\
            & SimLingo~\cite{Renz2025cvpr}& 85.07\makebox[0pt][l]{\pmsd{0.95}} & 67.27\makebox[0pt][l]{\pmsd{2.11}} \\ 
            & HiP-AD~\cite{tang2025hipadhierarchicalmultigranularityplanning}& 86.77 & 69.09  \\
            \midrule
            \multirow{2}{*}{\textbf{P}}
            & Think2Drive~\cite{li2024think2drive} & 91.85 & \textbf{85.41} \\
            & PlanT 2.0 (ours) & \textbf{92.4}\makebox[0pt][l]{\pmsd{1.7}} & 83.8\makebox[0pt][l]{\pmsd{3.3}} \\
            \midrule
            \textit{P} & \textit{Expert~\cite{sima2023drivelm}}&
            \textit{97.02}&
            \textit{92.27}
            \\
            \bottomrule
        \end{tabular}
    
    \caption{\textbf{Evaluation results on Bench2Drive.} \textbf{S} denotes sensor based models, \textbf{P} denotes privileged models. The scores of the expert are taken from \cite{zimmerlin2024tfpp}, the scores for Think2Drive are taken from \cite{yang2025raw2drivereinforcementlearningaligned}.}
    \label{tab:results_bench2drive}
\end{table}

Finally, we evaluate our model on Longest6 v2, the results are shown in Table~\ref{tab:results_longest6}. Our model reaches state-of-the-art performance on Longest6 v2, outperforming the reinforcement learning based CaRL and even the expert policy PDM-Lite. We are thus able to reproduce the result of the original PlanT, which reached expert-level performance on the Leaderboard 1.0 version of Longest6, but we fail to reach the same level of performance under the newly introduced Leaderboard 2.0 scenarios, highlighting their level of difficulty. In the following, we investigate this performance gap through a detailed analysis of our final model's failures and learned shortcuts.

\begin{table}[t]
    \centering
    \setlength{\tabcolsep}{0.015\textwidth}
        \begin{tabular}{l|cccc}
            \toprule
            \textbf{Method} &
            \multicolumn{1}{c}{\textbf{DS} $\uparrow$} &
            \multicolumn{1}{c}{\textbf{RC} $\uparrow$} &
            \multicolumn{1}{c}{ \textbf{CP} $\downarrow$} \\
            \midrule
          
            CaRL~\cite{JaegerCarl2025} &64\pmsd{2}&82\pmsd{1} & \textbf{0.01}\\
            PlanT 2.0 (ours) & \textbf{74.7}\pmsd{1.1} & \textbf{98.8}\pmsd{1.0} & 0.03 \\
            \midrule
            \textit{Expert~\cite{sima2023drivelm}}&\textit{73}&\textit{100} & 0.00\\
            \bottomrule
        \end{tabular}
    \caption{\textbf{Evaluation results on Longest6 v2.} All presented approaches are privileged planners. The scores of the expert are taken from \cite{JaegerCarl2025}}
    \label{tab:results_longest6}
\end{table}

\boldparagraph{Ablation studies}
 To select our planning representation and generation method, we perform a rigorous evaluation of the nine different combinations of our three proposed representations and generation methods. The detailed results of these evaluations can be found in Section~\ref{sec:output_representation}.  We observe significant performance differences between the different methods, showing the importance of intentionally choosing between different planning approaches. 

 We find that using a linear layer for waypoint generation generally outperforms the more complex GRU based approaches, while the planning representation using both waypoints and path points outperforms other approaches on average. The model combining these two methods achieves the highest score overall and represents our final model.
\section{Model Analysis}

In this chapter, we analyze our final model, PlanT 2.0, by investigating failure modes on the validation routes and exposing learned shortcuts through the evaluation of the model on perturbed inputs.

\subsection{Lack of environmental understanding}
\begin{figure}[t]
    \centering
    \includegraphics[trim={20mm 5mm 20mm 10mm}, clip, height=3.5cm]{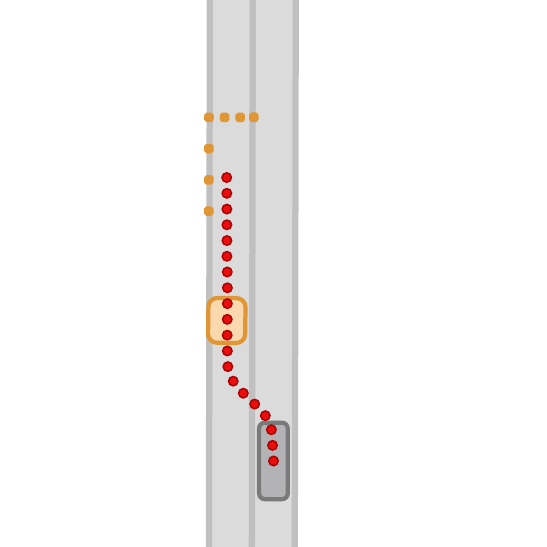}
    \hfill
    \includegraphics[trim={20mm 5mm 20mm 10mm}, clip, height=3.5cm]{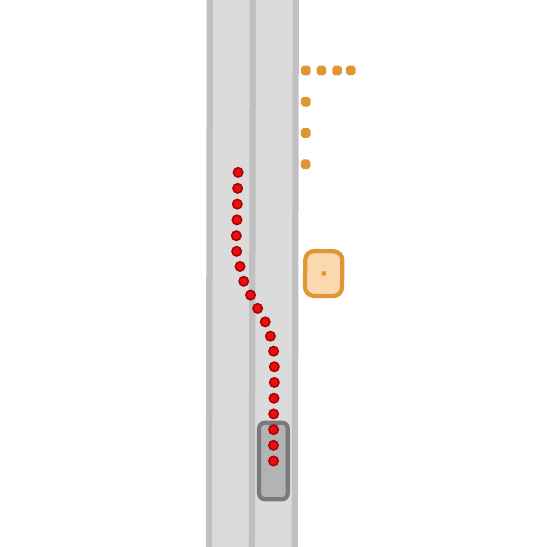}
    \hfill
    \includegraphics[trim={20mm 5mm 20mm 10mm}, clip, height=3.5cm]{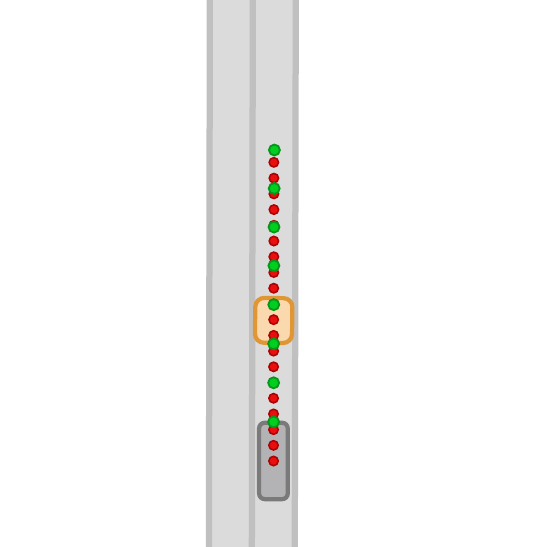}
    \legend{true}{false}{false}{true}{true}{true}
    \caption{\textbf{Lack of environmental understanding.} Incorrect reactions to different permutations of the construction obstacle. The first image shows the path avoiding an obstacle in the ego lane, while the obstacle is in the left lane. The second image shows the model adjusting the path for an obstacle next to the road, the third example shows the model failing to react to the obstacle without the presence of construction cones.}
    \label{fig:construction_overfit}
\end{figure}

By manually altering a construction obstacle scenario, we expose a major flaw in the model's environmental understanding. Figure~\ref{fig:construction_overfit} shows three examples: (left) moving the construction to the opposite lane, (middle) moving it to the right side of the ego lane, and (right) removing the cones behind the construction sign.
Instead of forming an understanding about the environment and the role of static obstacles, it learns a defined reaction that gets triggered when the construction obstacle scenario is present, independent of the location. 
As shown in Figure~\ref{fig:construction_overfit}, when moving the obstacle to the opposite lane or to the right of the ego lane, the model still reacts as if the obstacle appeared in the ego lane. Furthermore, we find that the model specifically reacts to the cones behind every construction, and completely ignores the main traffic warning, as can be seen in the rightmost image of Figure~\ref{fig:construction_overfit}. This shows that the lack of variance in CARLA obstacles allow the model to learn unreasonable shortcuts, leading to policies that do not generalize. This effect could likely be reduced by including a wider range on obstacles during training, forcing the model to learn a geometric understanding of obstacles in the environment.

\insight{\textbf{Insight 1:} The absence of obstacle diversity leads to a lack of spatial reasoning, requiring the implementation of novel obstacle scenarios in CARLA.} %

\subsection{Trajectory generalization}
\begin{figure}[t]
    \centering
    \includegraphics[trim={20mm 5mm 20mm 10mm}, clip, height=3.5cm]{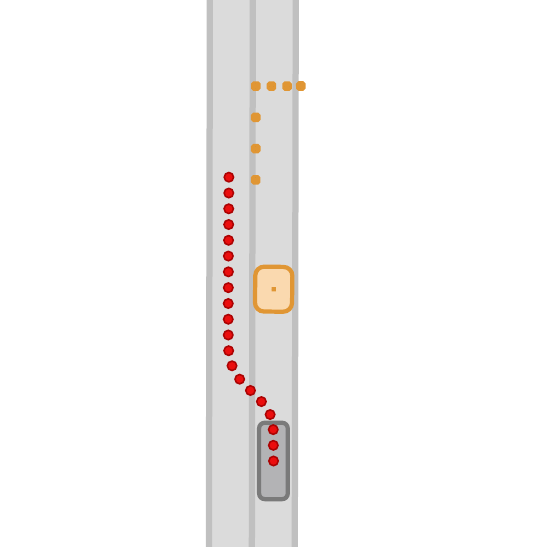}
    \hfill
    \includegraphics[trim={20mm 5mm 20mm 10mm}, clip, height=3.5cm]{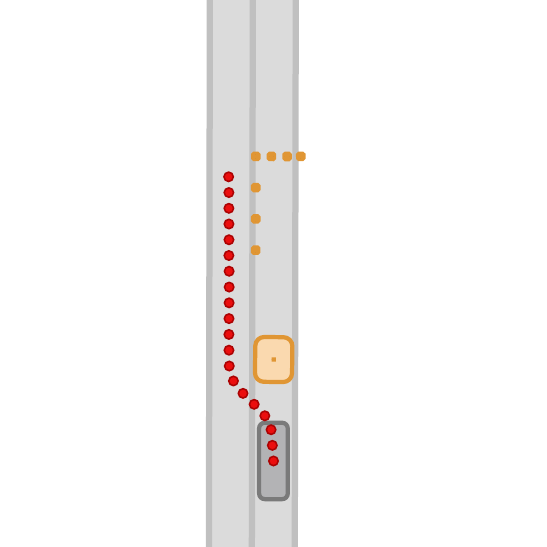}
    \hfill
    \includegraphics[trim={20mm 5mm 20mm 10mm}, clip, height=3.5cm]{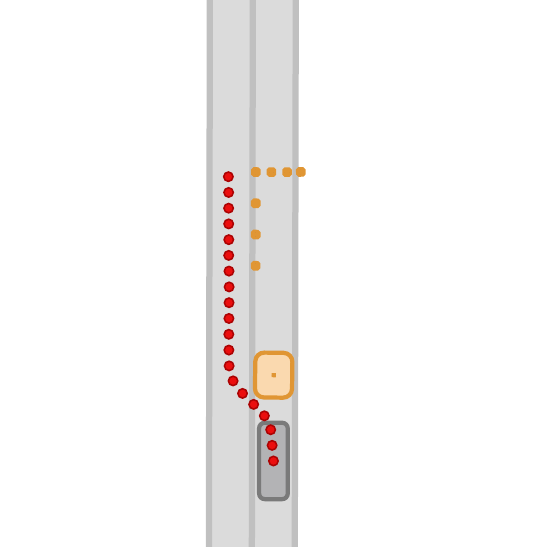}
    \legend{true}{false}{false}{true}{false}{true}
    \caption{\textbf{Trajectory generalization.} Predicted path at different distances to a construction obstacle. On the left, the model plans the correct path. In the other two images, as the vehicle is moved closer to the obstacle, the model is unable to adjust the steepness of the lane transition, resulting in trajectories that cause collisions. }
    \label{fig:construction_translation}
\end{figure}

Real-world driving situations require a wide variety of behaviours to correctly react to them. For example, when changing lanes to avoid an obstacle, the planned trajectory (i.e., the steepness of the lane transition) strongly depends on the ego vehicle’s speed and its distance to the obstacle. We analyse PlanT 2.0 in such situations and show the results in Figure~\ref{fig:construction_translation}. We vary the ego vehicle's distance from an obstacle and analyze the resulting predicted trajectories. The model performs reliably at moderate distances but fails to adjust its path when positioned too close, producing trajectories that collide with the obstacle. 
When looking at the dataset, it exhibits very low variance in lane transitions, since the expert policy consistently produces the same trajectories, and only a limited set of scenarios involves lane changes. As a result, the model primarily memorizes the trajectories generated by the expert and fails when alternative lane transitions are required. 
\insight{\textbf{Insight 2:} Fixed expert trajectories lead to overfitting and failure to generalize lane transitions needed for obstacle avoidance. Datasets need more variance in the expert trajectories to prevent memorization of trajectories and improve generalization.}

\subsection{Proximity violations in planned trajectories}
\begin{figure}[t]
    \centering
    \includegraphics[trim={25mm 10mm 30mm 30mm}, clip, height=3.3cm]{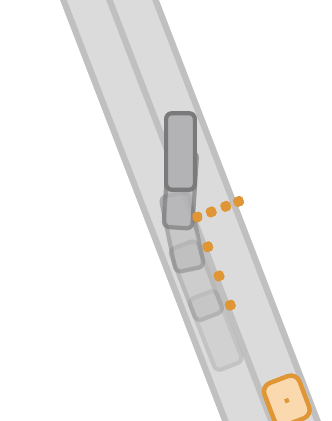}
    \hfill
    \includegraphics[trim={30mm 10mm 30mm 30mm}, clip, height=3.3cm]{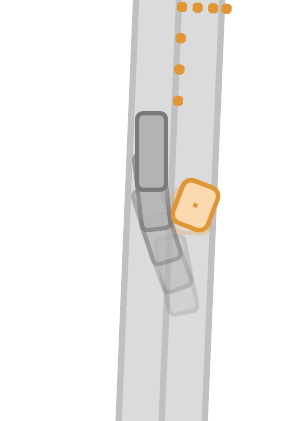}
    \hfill
    \includegraphics[trim={25mm 10mm 25mm 30mm}, clip, height=3.3cm]{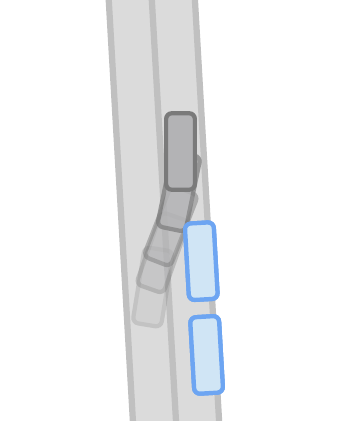}
    \legend{true}{true}{false}{true}{false}{false}
    \vspace{-1em}
    \caption{\textbf{Proximity violations in planned trajectories.} Three examples of side-swipe collisions in obstacle avoidance scenarios}
    \label{fig:obstacleavoidance}
\end{figure}

During obstacle avoidance, failures occasionally occur where the model plans a path around an obstacle but leaves insufficient clearance, causing the vehicle to graze it. As shown in Figure~\ref{fig:obstacleavoidance}, this most often occurs when the model returns to its lane too early. Two potential reasons for these failures could be either poor controller execution or inaccurate trajectory prediction. Since we use the same controller as the expert model, which does not encounter this issue, the problem lies in the predicted trajectory. 
Our expert, PDM-Lite, plans trajectories very close to obstacles; even small deviations can lead to collisions. Failures arise from inference uncertainty combined with the expert trajectories’ minimal margins for error. 
\insight{\textbf{Insight 3:} Small margins in the expert demonstrations cause failures under uncertainty. Introducing larger safety margins could mitigate this, though at the cost of less efficient obstacle avoidance.}

\subsection{Positional shortcuts}
\begin{figure}[t]
    \includegraphics[trim={20mm 6mm 25mm 12mm}, clip, height=3.3cm]{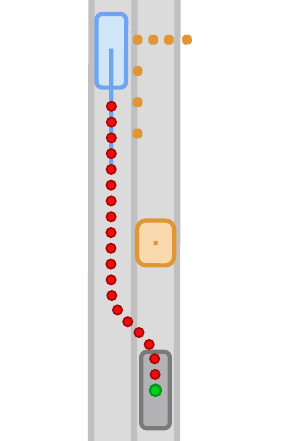}
    \hfill
    \includegraphics[trim={25mm 6mm 25mm 12mm}, clip, height=3.3cm]{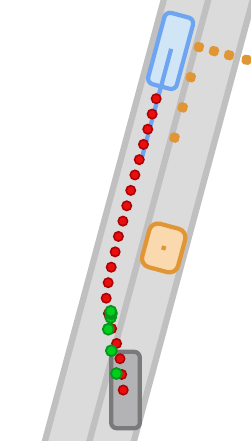}
    \hfill
    \includegraphics[trim={15mm 6mm 0mm 12mm}, clip, height=3.3cm]{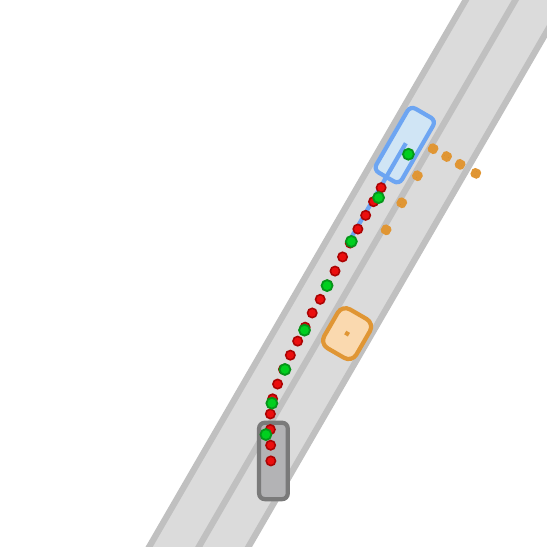}
    \legend{true}{true}{false}{true}{true}{true}
    \vspace{-1em}
    \caption{\textbf{Positional shortcuts.} Predicted path and waypoints in a construction obstacle scenario with oncoming traffic at manually altered rotations of 0, 15 and 30 degrees. As the rotation increases, the model predicts higher driving speeds.}
    \vspace{-0.5em}
    \label{fig:construction_rotation}
\end{figure}

\begin{figure}[t]
    \centering
    \includegraphics[width=\linewidth]{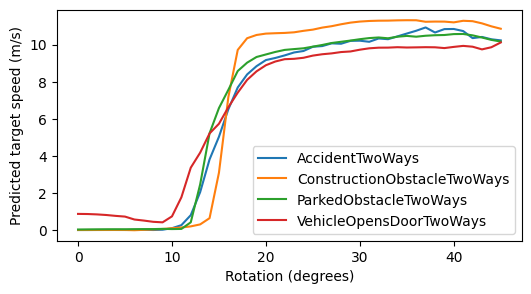}
    \vspace{-1em}
    \caption{Predicted target speeds for different ego vehicle rotations across four obstacle scenarios with oncoming traffic preventing the overtaking maneuver.}
    \label{fig:rotation_plot}
\end{figure}

A robust model needs to be able to adaptively react to situations and reconsider a decision.
We find that the model learns a shortcut by using its own rotation as a predictive signal for when the road is clear to pass an obstacle. Once a decision is made to go around the obstacle, it ignores the scene and does not reconsider the decision. To quantify this, we systematically rotate the ego vehicle in obstacle avoidance scenarios and measure how the model's target speed prediction varies with orientation. 
An example can be seen in Figure~\ref{fig:construction_rotation}, where the model predicts to continue driving as its rotation increases. 
Figure~\ref{fig:rotation_plot} plots predicted target speed against ego rotation across multiple obstacle avoidance scenarios with oncoming traffic. In all four scenarios, at rotation values around 10-15 degrees, the predicted speed abruptly increases to signal regular driving. \\
This arises because the expert PDM-Lite only initiates a turn when the road is free. Since the dataset includes only correctly solved samples, the model learns to always complete an overtaking maneuver once initiated, ignoring potential traffic.
Although positional augmentation during training mitigates this effect, the model continues to be influenced by high rotation values.
\insight{\textbf{Insight 4:} Shortcuts introduced by rigid expert demonstrations can be exploited and lead to unrealistic behavior. Generalization can be improved by covering more diverse driving situations or more aggressive augmentation.}

\subsection{Scenario anticipation}
\label{sec:scenario_anticipation}
\begin{figure}[t]
    \centering
    \includegraphics[trim={25mm 1mm 25mm 25mm}, clip, height=3.5cm]{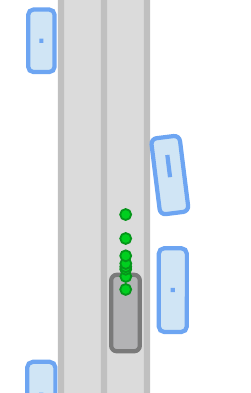}
    \hfill
    \includegraphics[trim={25mm 1mm 25mm 25mm}, clip, height=3.5cm]{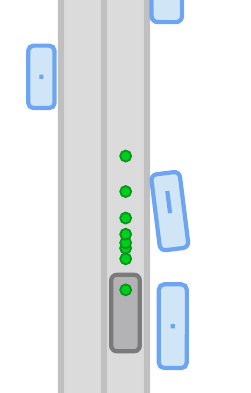}
    \hfill
    \includegraphics[trim={25mm 1mm 25mm 25mm}, clip, height=3.5cm]{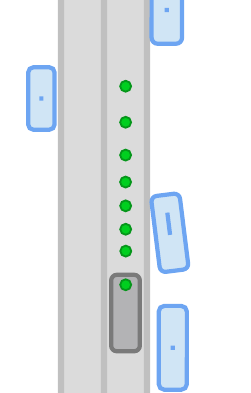}
    \legend{true}{true}{false}{false}{true}{false}
    \caption{\textbf{Scenario anticipation.} Predicted waypoints at different distances to a parked vehicle that cuts into the ego lane. In the first image, the model correctly predicts a slight deceleration to let the parked vehicle merge and then to continue driving. The second image shows that the model correctly predicts waiting until the parked vehicle has merged. In the third image, the model incorrectly predicts driving at 36 km/h.}
    \label{fig:parkingcutin_fail}
\end{figure}

\begin{figure}[t]
    \centering
    \includegraphics[width=\linewidth]{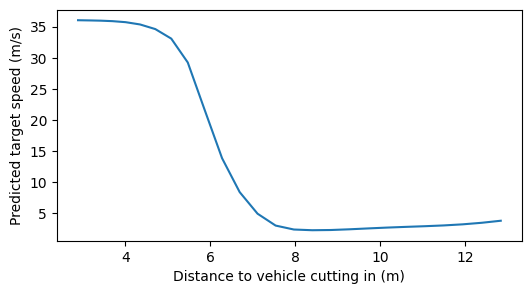}
    \caption{Predicted target speed plotted against distance to the vehicle cutting in. At x-values larger than 6, we observe the expected behaviour of driving slower with decreasing distance to the parked vehicle. Below 6m, the model incorrectly predicts regular driving at 36 km/h.}
    \label{fig:parkingcutin_plot}
\end{figure}

Another failure arises from the interaction between the model and CARLA’s scenario mechanics. In the ParkingCutIn scenario, a parked vehicle cuts in front of the ego vehicle, forcing it to brake. Since our waypoint representation extends two seconds into the future, the model learns to anticipate the event even before the parked vehicle moves. 
We analyze this systematically by manually altering the ego position in one sample (see Figure~\ref{fig:parkingcutin_fail}). As shown in Figure~\ref{fig:parkingcutin_plot}, when the ego is shifted closer to the cutting-in vehicle, the model predicts acceleration at distances below 6 meters, ignoring the parked vehicle and causing a collision.
The model picks up spurious cues in the scene or even \textit{hallucinates} the reason for braking, which can result in random braking maneuvers.
Although this is a conceptual issue, CARLA amplifies it because the arrangement of parked cars in the scenario is unique and can be recognized in advance. Similar anticipatory behaviors appear in other parked-vehicle scenarios, but in this scenario, they lead to a collision. CARLA triggers the parked vehicle based on the ego vehicle’s velocity. When the model brakes too early, the trigger is delayed, producing an out-of-distribution situation. The model then accelerates incorrectly, as it has never encountered a late cut-in during training. Figure~\ref{fig:parkingcutin_fail} illustrates this: the two left images show the model predicting a reduced speed as the ego approaches the parked vehicle, while the right image shows the model predicting full-speed driving even though the ego vehicle could already be stopped, since it does not observe ego velocity.
 
\insight{\textbf{Insight 5:} Lack of variety in CARLA scenario timings and expert behaviors leads to problematic emergent behaviors. This could be mitigated by including data from different distributions, such as modified CARLA scenarios or real-world data, and by introducing variance into the expert policy.}

\subsection{Inherent modeling of vehicle dynamics}
The waypoint representation inherently encodes vehicle dynamics, which can cause problems in scenarios requiring large acceleration or deceleration. Since target speeds derived from waypoints reflect the predicted speed rather than the desired speed, the model may not achieve appropriate acceleration. This could be mitigated by using a more advanced control method, such as model predictive control, to adjust acceleration according to the predicted trajectory. However, if a driving system is deployed to different vehicles, the learned dynamics may no longer match the actual vehicle, for example by predicting acceleration values only possible in some vehicles. %

\begin{figure}[t]
    \begin{subfigure}[t]{0.3\linewidth}
        \centering
        \includegraphics[trim={70mm 70mm 70mm 50mm}, clip, height=2.5cm]{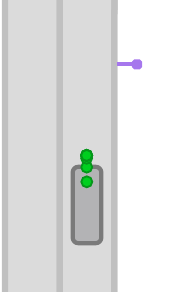}
        \\
        \hspace{0.01mm}
        \includegraphics[height=8mm]{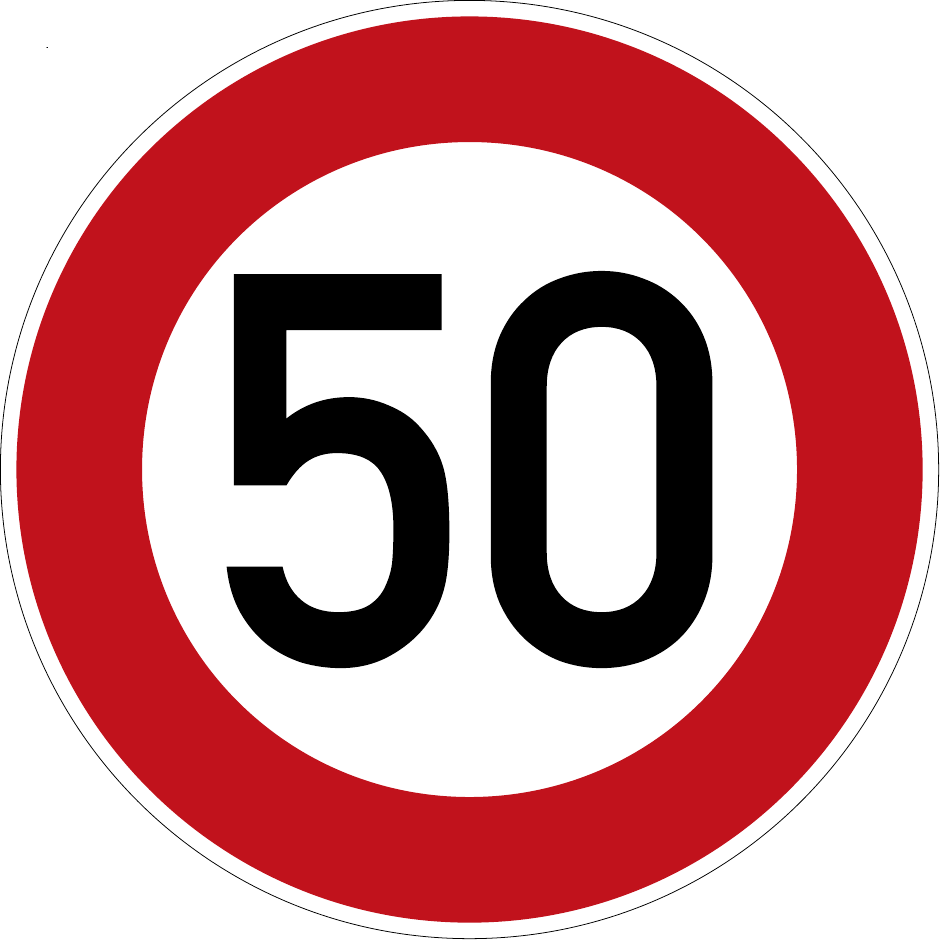}\hspace{5.3mm}
    \end{subfigure}
    \hfill
    \begin{subfigure}[t]{0.3\linewidth}
        \centering  
        \includegraphics[trim={70mm 70mm 70mm 50mm}, clip, height=2.5cm]{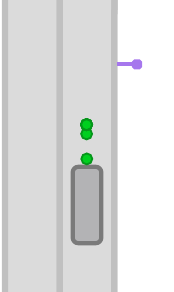}
        \\
        \hspace{0.01mm}
        \includegraphics[height=8mm]{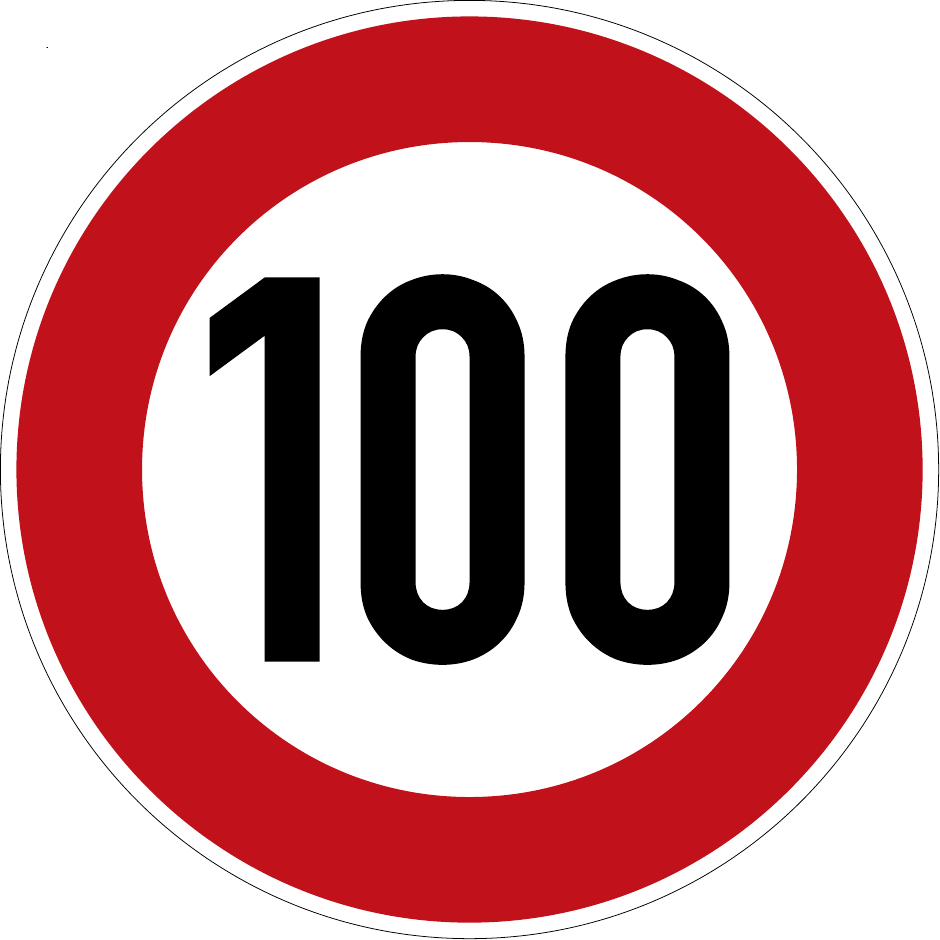}\hspace{5.3mm}
    \end{subfigure}
    \hfill
    \begin{subfigure}[t]{0.3\linewidth}
        \centering
        \includegraphics[trim={70mm 70mm 70mm 50mm}, clip, height=2.5cm]{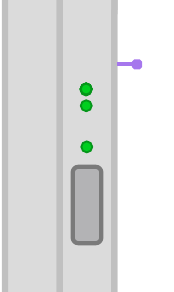}
        \hspace{0.01mm}
        \\
        \includegraphics[height=8mm]{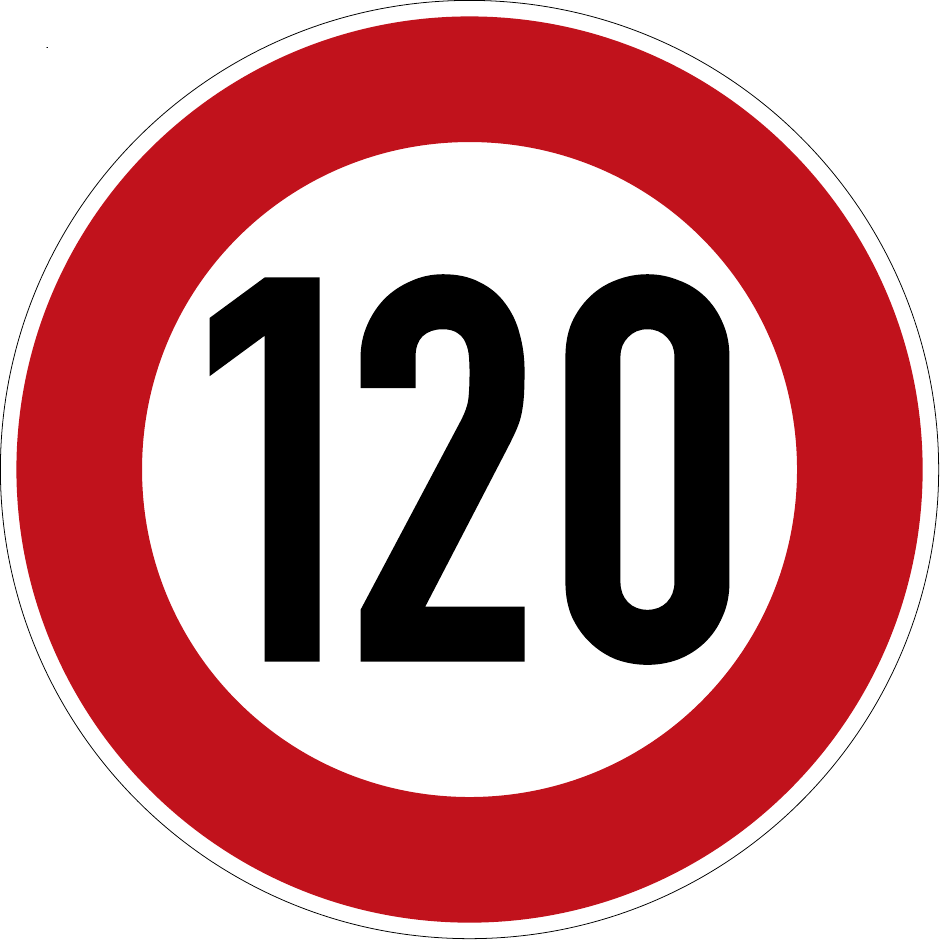}\hspace{5.3mm}
    \end{subfigure}
    \\
    \legend{true}{false}{true}{false}{true}{false}
    \caption{\textbf{Inherent modeling of vehicle dynamics.} The model's reaction to a pedestrian crossing at differing speed limits (50, 100, 120 km/h). With increasing speed limits, the model predicts an increased braking distance.}
    \label{fig:pedestrianbrake}
\end{figure}

We observe another problematic behavior that can not be solved by improved control methods. We analyze the model's braking behavior under varying speed limits in a pedestrian scenario. Figure~\ref{fig:pedestrianbrake} shows that as the speed limit increases, the predicted stopping point shifts closer to the pedestrian, suggesting that the model infers higher speed limits correspond to higher ego velocities and longer braking distances. Consequently, if the agent is already moving slowly, it may stop decelerating or even accelerate to match a familiar braking trajectory, increasing collision risk. This could be addressed by introducing an additional classifier for hazard braking or by including the ego vehicle's current velocity in the input. During preliminary experiments, the benefits of including the current velocity in our model were outweighed by performance drops due to causal confusion, in line with findings of other works \cite{Codevilla2019ICCV,Chitta2023PAMI,chen2023end}.

\insight{\textbf{Insight 6:} Waypoint-based planning embeds vehicle dynamics, causing a mismatch between predicted and desired speeds that can lead to unsafe behavior. This could be solved by reducing the dependency on vehicle dynamics, or carefully including them in the model.}

\subsection{Undefined pre- and post-crash behavior} 
\label{sec:crash}
\begin{figure}[t]
    \centering
    \includegraphics[trim={60mm 65mm 65mm 15mm}, clip, height=3.5cm]{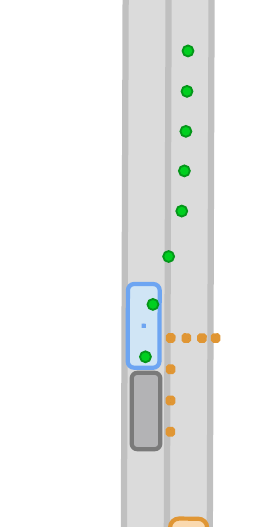}
    \hfill
    \includegraphics[trim={40mm 65mm 60mm 15mm}, clip, height=3.5cm]{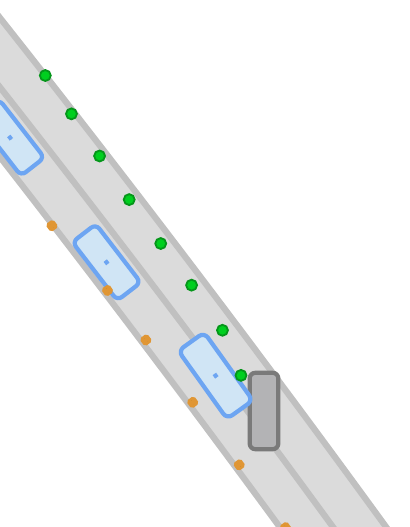}
    \hfill
    \includegraphics[trim={40mm 65mm 40mm 15mm}, clip, height=3.5cm]{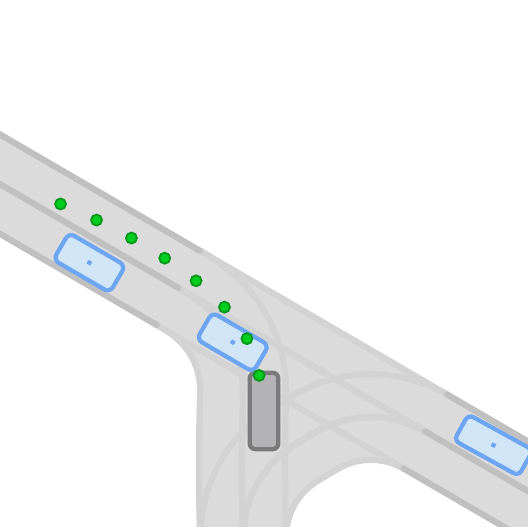}
    \\

\legend{true}{true}{false}{true}{true}{false}

    \caption{\textbf{Undefined pre- and post-crash behavior.} Three post-crash scenarios, showing the model predicting to continue driving.}
    \label{fig:crashes}
\end{figure}

Since the dataset contains only successful driving samples, the model has no reference for handling imminent collisions or post-collision behavior. Consequently, it defaults to regular driving in such situations, increasing the risk of further collisions and higher penalties. During evaluation, we observe numerous cases where the model continues to accelerate after a collision, often causing additional impacts. Figure~\ref{fig:crashes} illustrates three post-crash scenarios where the model attempts to keep driving.

Although unrealistic, this behavior can sometimes be beneficial in CARLA: continued acceleration may allow the vehicle to exit the scene and resume its route, improving the overall score despite accumulating collision penalties. Alternatively, the model could remain idle until the scenario times out, despawning other actors and incurring a smaller penalty. However, this approach risks triggering a “vehicle blocked” infraction, which can terminate the route entirely.

Future work could explore how different post-crash strategies affect performance and investigate ways to intentionally include appropriate reactions in training data, without artificially introducing collisions.
\insight{\textbf{Insight 7:} Correct behavior in (near-)collision situations is not specified by the dataset and needs to be carefully considered to optimize for performance metrics or realism.}

\subsection{Sparse environment representation}
A notable failure occurs in the InvadingTurn scenario, where oncoming vehicles slightly encroach on the ego lane, requiring the ego to yield. In some rural scenarios, fences along the right side of the road limit the vehicle’s ability to maneuver. The expert policy accounts for these fences by reducing the space allocated to oncoming traffic. Our model, however, does not receive fence information as input and cannot adjust appropriately, resulting in repeated collisions along the fence.

This illustrates a key limitation: incorporating map elements or minor environmental details is difficult because the transformer's computational cost grows quadratically with the number of input objects. The BEV SD-Map has a resolution of 50cm, which cannot capture such fine details accurately. Increasing map resolution would further raise computational costs and conflict with the sparse input design.

\insight{\textbf{Insight 8:} Sparse input representations prevent the model from accounting for surrounding environmental details, calling for other methods of specifying occupancy in the environment.}

\section{Conclusions}

In this work, we have identified multiple limitations of training autonomous driving systems in CARLA, highlighting the impact of the expert behavior and a lack of variance in CARLA scenarios and expert demonstrations. We revisit PlanT and upgrade it to solve the challenges posed by the CARLA leaderboard 2.0. Our improved model achieves state-of-the-art performance across multiple benchmarks, and even reached expert performance on the Longest6 v2 benchmark, while being reproducible and open source. We have additionally analysed the limitations of this model and suggest a wide range of improvements for further research.

\boldparagraph{Limitations} 
We perform our analysis in simulation and rely on privileged inputs, leaving questions about how well the findings transfer to sensor-based methods. However, since most of our findings are caused by the underlying dataset and expert policy, our conclusions can likely be transferred to sensor-based models in CARLA. Second, some of the identified failure modes may be specific to CARLA’s scenario design or to the expert policy used for data collection, which may limit the application of our findings beyond CARLA research. 

Our final model fails to reach expert performance on the validation routes, further research could focus on implementing our proposed improvements and investigating their effects. Additionally, training without Town13 greatly reduces the model's performance, showing a lack of generalization and a domain gap between the validation routes and the training routes excluding Town13. We leave the development of a more comprehensive set of training routes to future research.

\vspace{0.2cm}
\noindent \textbf{Acknowledgements.}
This project was supported by the DFG EXC number 2064/1 - project number 390727645 and by the German Federal Ministry for Economic Affairs and Climate Action within the project “NXT GEN AI METHODS – Generative Methoden für Perzeption, Prädiktion und Planung". We thank the International Max Planck Research School
for Intelligent Systems (IMPRS-IS) for supporting K. Renz.

\clearpage
{
    \small
    \bibliographystyle{ieeenat_fullname}
    \bibliography{bib/bibliography_org,bib/final_bibliography_dblp,bib/bibliography_long}
}

\clearpage
\appendix
\clearpage
\maketitlesupplementary

\section{Implementation Details} \label{supp:implementation}
We make minor modifications to the data collection to fix bugs and improve the expert's realism. We implement a fix which replaces incorrect extent values in the dataset, as CARLA reports incorrect extent values for static objects. Additionally, since CARLA does not give penalties for vehicle collisions if the ego vehicle is stationary, we manually filter samples where such a collision has occured. Lastly, we introduce a small modification to the expert behavior in the scenarios OppositeVehicleRunningRedLight and OppositeVehicleTakingPriority, where an emergency vehicle crosses an intersection without the right of way. In some cases, the expert recognizes that it is able to pass the intersection before the emergency vehicle has arrived, which is an undesired behavior. We prevent this by introducing additional heuristics during these scenarios.

To support the model's ability to recover from route deviations, we include the same data augmentation strategy as used by \cite{zimmerlin2024tfpp}. This introduces samples where the ego position is translated and rotated by a small amount, simulating a recovery case for the model to learn from, reducing the impact of distribution shifts during evaluation.

\section{Model improvements} \label{supp:model}

\boldparagraph{Input representation}
In order to obtain a driving model capable of real-world driving, we extend the object representation of PlanT to include five object classes in addition to vehicles. %
All objects are represented by an oriented bounding box and a velocity scalar and are encoded by a class-specific linear projection.\\
\textit{(1) Pedestrians:} We include pedestrians only when they are moving to prevent a deadlock where the model waits for a pedestrian to cross, while the pedestrian scenario is still waiting for the vehicle to come closer. \\
\textit{(2) Static Objects:} Static objects are only included from construction scenarios, as these are the only static obstacles currently relevant in CARLA. \\
\textit{(3) Emergency Vehicles:} Since emergency vehicles often require a different reaction to standard vehicles and because PlanT is not able to distinguish between emergency and regular vehicles visually, we use a different object class for emergency vehicles. \\
\textit{(4) \& (5) Stop Signs and Traffic Lights:} PlanT previously ignored stop signs completely and input the presence of a red light through a binary flag in the waypoint GRU. %
Since this introduces ambiguities about the correct position to stop in, we include stop signs and traffic lights in the object representation using bounding boxes that indicate positions the vehicle must not cross while a red traffic light is active or where it must come to a stop for a stop sign. They are only included when relevant to the ego vehicle, specifically, when the vehicle is affected by a red or yellow traffic light, or when it has not yet come to a complete stop at a stop sign.

We remove the route information from the object representation and replace it with a single input token by encoding 20 route points in BEV coordinate space sampled at distances of 1 meter using a linear layer. Previously, two oriented bounding boxes were input for route information, which could lead to a loss of information in turns, where the bounding boxes become shorter to represent the curved road. %

Additionally, since PlanT can't determine the current speed limit visually, we include this information by using a learned token with four discrete states, corresponding to the four possible speed limit values in CARLA. Although exceeding the speed limit does not result in infractions in CARLA, incorporating this information reduces uncertainty in waypoint spacing.

\boldparagraph{Input range} 
To allow PlanT 2.0 to perform longer distance planning, especially at high speeds and in the presence of fast-moving actors, we increase the spatial object detection range from the previous radius of 30m. For scenarios where the ego vehicle has to cross into oncoming traffic, a very high lookahead distance of up to 100m is required. Since using a radius of 100m would input many irrelevant objects, we instead use a radius of 50m for objects behind the ego vehicle ($x < 0$), and an ellipse shape with a radius of 100m in the x-direction and 50m in the y-direction for objects in front of the ego vehicle ($x > 0$).

To account for faster driving speeds and higher scenario difficulty, we increase the object detection range from the previous radius of 30m. We use a circular radius of 50m for objects behind the ego vehicle, and an ellipse with an extent of 100m in the x-axis and 50m in the y-axis for objects in front of the vehicle.\\

\boldparagraph{SD-map input} 
To successfully navigate complex environments where swerving and lane change behaviours are necessary, some basic information about the road layout is crucial. 
To address this, we include a BEV representation of the surrounding road geometry (i.e., SD map) in the input. 
We generate a 128×128 pixel RGB image representing a 64×64 meter area centered around the ego vehicle. The SD-map encodes four classes: background, drivable road, solid lane markings, and broken lane markings. This representation allows the model to distinguish between lanes of the same and opposite direction, as well as to infer the number and type of lanes available. The SD-map is processed through a ResNet-18 encoder, followed by a linear layer to produce a single input token. 

\subsection{Waypoint representation}
The representation of the planning output plays a crucial role in the performance and robustness of autonomous driving systems. While traditional models typically predict a fixed number of future waypoints at regular time intervals, recent research has shown that the structure of this output can significantly influence downstream driving behavior \cite{Jaeger2023ICCV}. %

We use the linear regression controller implemented by \cite{Beißwenger2024PdmLite} for longitudinal control, which determines throttle and braking values from the current and desired speed. For waypoint methods, the target speed is obtained from the distance between the third and fourth waypoints. 
For lateral control, we again adopt the control method of the expert, which is based on a PID controller using a variable lookahead distance. 
We propose three different waypoint representations to investigate the effect of decoupled lateral and longitudinal control, as well as the method used for predicting the longitudinal control. %

\boldparagraph{Waypoints (WPS)} The baseline model predicts eight waypoints at 4 Hz, used for both lateral and longitudinal control. 

\boldparagraph{Disentangled Route + Speed Classifier (PATH)} Following the approach of \cite{Jaeger2023ICCV}, this variant predicts the route as 20 spatially equidistant points for lateral control, while estimating target speed through a two-hot encoded classifier. This disentangled design has been shown to reduce collisions with static objects, likely because it provides the route prediction with consistent training signals throughout training, including at standstill, where traditional waypoint-based signals collapse. Moreover, separating lateral and longitudinal control helps prevent ambiguity in one mode from influencing the other. We adopt the same speed classifier used in \cite{Jaeger2023ICCV}.

\boldparagraph{Disentangled Route + Waypoints (P+WP)} \cite{Renz2025cvpr} combines spatial route points for lateral control with traditional temporal waypoints for longitudinal control. Temporal waypoints provide richer information about the vehicle's future trajectory compared to a single target velocity. We use 20 route points for steering and 8 waypoints for speed control, respectively.

\begin{figure}[t]
    \centering
    \includegraphics[trim={70mm 68mm 75mm 22mm}, clip, height=3cm]{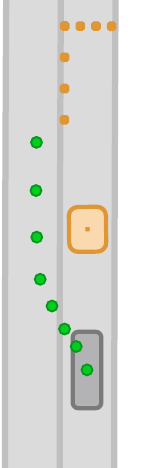}
    \hfill
    \vrule 
    \hfill
    \includegraphics[trim={32mm 55mm 26mm 35mm}, clip, height=3cm]{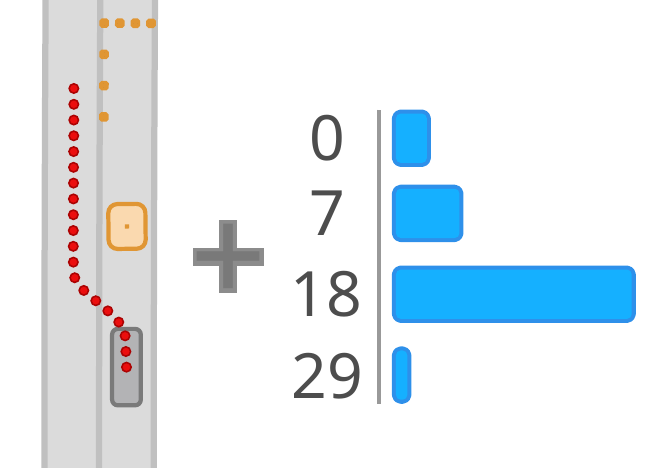}
    \hfill
    \vrule 
    \hfill
    \includegraphics[trim={32mm 55mm 63mm 35mm}, clip, height=3cm]{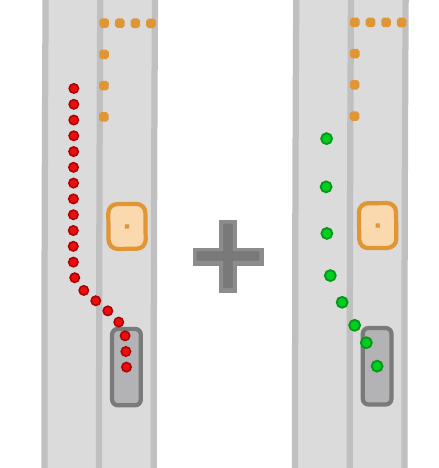}
    \caption{\textbf{Waypoint representation.} Visualizations for the three waypoint representations, from left to right: Only waypoints (WPS), Path and Speed classifier (PATH), Path and Waypoints (P+WP)}
    \label{fig:wp_rep}
\end{figure}

An illustration of the different waypoint representations is shown in Figure~\ref{fig:wp_rep}.

\subsection{Waypoint generation}
\begin{figure}[t]
    \centering
    \includegraphics[width=1\linewidth]{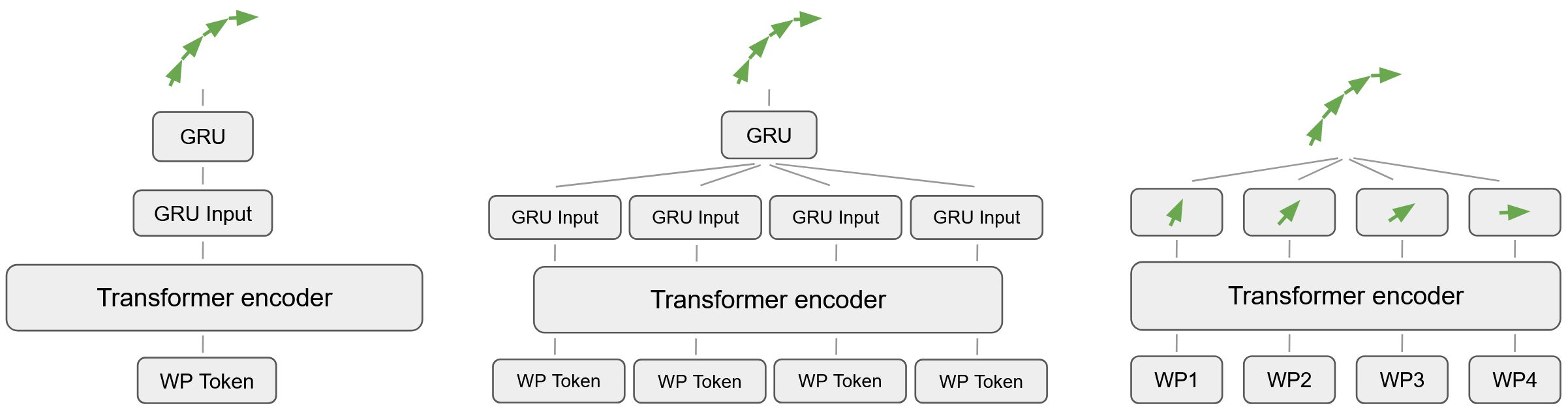}
    \caption{\textbf{Waypoint generation.} Visualizations of the three waypoint generation options, from left to right: Single Token GRU, Multi Token GRU, Multi Token Linear}
    \label{fig:wp_generators}
\end{figure}

Different implementations of learning-based planners employ varying methods for generating waypoints, often making it unclear whether improvements stem from the waypoint representation itself or simply from the method of generation. We examine the impact of the waypoint generation method by performing a comparison of 3 different approaches, which are visualized in Figure~\ref{fig:wp_generators}.

\boldparagraph{Single-token GRU (Baseline)}
The baseline implementation uses a GRUCell where a single token from the transformer is used to initialize the hidden state. Waypoints are then generated autoregressively: at each step, the GRU takes as input the previously predicted waypoint concatenated with the target point. This setup limits the model to a single transformer token for all predictions, relying on the GRU's hidden state to capture temporal dynamics. This is the method used in the original PlanT implementation.

\boldparagraph{Multi-token GRU}
\cite{zimmerlin2024tfpp} employs a GRU that receives a full input sequence of transformer-encoded tokens, one for each waypoint, modeling inter-waypoint dependencies more explicitly. The target point is encoded into the GRU's initial hidden state rather than being provided at each timestep.

\boldparagraph{Multi-token Linear}
\cite{Renz2025cvpr} utilizes a simplistic waypoint generation method, also encoding a token per output point, but regressing the x and y values using a simple linear layer. The final waypoints are obtained by cumulatively summing the points in order.

\section{Evaluation results}
\boldparagraph{Input Representation} 
\begin{table}[t]
    \setlength{\tabcolsep}{0.025\textwidth}
    \centering
    \begin{tabular}{l|cc}
    \toprule
     & \textbf{DS} $\uparrow$ & \textbf{SR} $\uparrow$ \\
    \midrule
    PlanT 2.0 (ours) & 92.4 \pmsd{1.7} & 83.8 \pmsd{3.3} \\
    w/o SD map & \textbf{93.3} \pmsd{3.3} & \textbf{84.9} \pmsd{4.5}  \\
    w/o higher range & 90.4 \pmsd{2.2} & 80.3 \pmsd{3.2}  \\
    \bottomrule
    \end{tabular}
    \caption{\textbf{Input representation ablations on Bench2Drive.}}
    \label{tab:input}
\end{table}

We perform an ablation of the new inputs on Bench2Drive by first removing the SD map and additionally returning the range to the previous setting. As can be seen in Table~\ref{tab:input}, the model performs slightly better on Bench2Drive without the SD map input, this is likely caused by CARLA's limited diversity and the scenarios often being solvable without any environmental awareness. However, we choose to keep the SD map in the final model, as a model without any input of the surrounding road layout is unlikely to perform well in realistic traffic situations. As expected, decreasing the range to previous values leads to a drop in performance on Bench2Drive.

\boldparagraph{Output Representation} 
\label{sec:output_representation}
We evaluate each of the three waypoint representations in combination with each of the three waypoint generation methods, leading to 9 different configurations. For every configuration, we train three models using different training seeds and evaluate every model with three evaluation seeds.

\begin{table}[t]
    \setlength{\tabcolsep}{0.007\textwidth}
    \renewcommand{\arraystretch}{1.2} %
    \centering
    \begin{tabular}{c|ccc|c}
    \toprule
     \textbf{Normalized DS} $\uparrow$ & WPS & PATH & P+WP & \textit{Mean}\\
     \midrule
     Single Token GRU & \phantom{0}7.4\pmsd{2.1} & 21.8\pmsd{2.2} & 20.0\pmsd{4.0} & \textit{16.4}\\
     Multi Token GRU & 11.3\pmsd{1.9} & 10.2\pmsd{4.3} & 24.0\pmsd{10.5} & \textit{15.2}\\
     Multi Token Linear & \phantom{0}8.1\pmsd{6.5} & 21.3\pmsd{8.3} & \textbf{28.6}\pmsd{2.9} & \textit{\textbf{19.3}}\\
     \midrule
     \textit{Mean} & \textit{8.9} & \textit{17.8} & \textbf{\textit{24.2}} & \textit{17.0}\\
     \bottomrule
    \end{tabular}
    \caption{\textbf{Results on CARLA validation routes for different output representations.} Driving Scores of all nine configurations, standard deviations are reported across training seeds.}
    \label{tab:3x3table}
\end{table}

Table~\ref{tab:3x3table} shows the Normalized Driving Scores (NDS) of the different planning configurations on the CARLA validation routes. The driving scores vary greatly between representations and generation methods, highlighting the importance of this often overlooked design choice. We observe that the waypoint generation methods using multiple tokens outperform the single token approach for the waypoint and path + waypoint representations. In contrast, the single token approach outperforms both multi-token methods using the path + speed classifier representation. %
Overall, the simple linear waypoint decoder outperforms both of the more complex GRU-based decoders.

\begin{table}[t]
    \centering
    \setlength{\tabcolsep}{0.010\textwidth} %
        \begin{tabular}{c|ccccc}
            \toprule
            \textbf{Method} &
            \multicolumn{1}{c}{\textbf{NDS} $\uparrow$} &
            \multicolumn{1}{c}{\textbf{RC} $\uparrow$} &
            \multicolumn{1}{c}{ \textbf{CV} $\downarrow$} &
            \multicolumn{1}{c}{ \textbf{CL} $\downarrow$} &
            \multicolumn{1}{c}{ \textbf{ST} $\downarrow$} \\
            \midrule
                WPS &  \phantom{0}8.1\pmsd{6.5} & 85.9\pmsd{8.2} & 0.95 & 0.27 & 0.16\\
                PATH & 21.3\pmsd{8.3} & 85.8\pmsd{5.6} & 0.59 & \textbf{0.10} & 0.05 \\
                P+WP & \textbf{28.6}\pmsd{2.9} & \textbf{90.6}\pmsd{3.9} & \textbf{0.40} & 0.14 & \textbf{0.04}\\
            \midrule
            \textit{Expert~\cite{sima2023drivelm}}&\textit{61.55}&\textit{92.35} & - & - & -\\
            \bottomrule
        \end{tabular}
    \caption{\textbf{Metric details for Multi Token Linear.} Detailed validation results of the different waypoint representations using the Multi Token Linear waypoint generation method.}
    \label{tab:linear}
\end{table}
To compare the different waypoint representations in more detail, we provide more detailed metrics for the evaluations using the multi-token linear waypoint generation method in Table~\ref{tab:linear}. 
Consistent with the findings of \cite{Jaeger2023ICCV}, we observe that both models using spatial route points for lateral control achieve higher driving scores and cause significantly fewer layout- and vehicle collisions as well as fewer scenario timeouts. This supports the hypothesis that providing a continuous and disentangled training signal for steering improves lateral control, allowing the model to navigate the environment more effectively. 
The representation that additionally uses waypoints for longitudinal control exhibits a further decrease in vehicle collisions and scenario timeouts. This gain likely stems from improved long-term planning capabilities, allowing the model to identify critical situations earlier and adjust its behavior more smoothly.

Among all methods, the combination of route points for steering and waypoints for longitudinal control, along with the linear waypoint generation method, achieves the best overall performance with a Driving Score of 28.6 and a Route Completion of 90.6. We use this model for our remaining ablations and final evaluations.

\end{document}